\pdfoutput=1
\PassOptionsToPackage{dvipsnames}{xcolor}
\documentclass[11pt]{article}

\usepackage[]{acl}

\usepackage{times}
\usepackage{latexsym}
\usepackage[T1]{fontenc}
\usepackage[utf8]{inputenc}

\usepackage{microtype}
\usepackage{amsmath,amssymb,amsfonts}
\usepackage{algcompatible}
\usepackage{algorithm}
\usepackage{algpseudocode}
\usepackage{cite}
\usepackage{hyperref}

\usepackage{graphicx}
\usepackage{textcomp}
\usepackage{xcolor}
\def\BibTeX{{\rm B\kern-.05em{\sc i\kern-.025em b}\kern-.08em
    T\kern-.1667em\lower.7ex\hbox{E}\kern-.125emX}}

\usepackage{multirow}
\usepackage{array}
\usepackage{nicematrix}
\usepackage{booktabs}%
\usepackage{enumitem}
\usepackage{tabularx}
\usepackage{amsmath, amssymb}
\usepackage{arydshln}
\usepackage{inconsolata}
\usepackage{soul}
\usepackage{makecell}
\usepackage{listings}
\usepackage{caption,subcaption}
\usepackage{soul}

\usepackage[author={An}]{pdfcomment}

\newcolumntype{?}{!{\vrule width 1pt}}

\graphicspath{ {./images/} }

\title{IgnitionInnovators at "Discharge Me!": Chain-of-Thought Instruction Finetuning Large Language Models for Discharge Summaries}

\author{
    An Quang Tang \\
    RMIT University, Australia\\
    s3695273@rmit.edu.vn
    \And
    Xiuzhen Zhang \\
    RMIT University, Australia\\
    xiuzhen.zhang@rmit.edu.au
    \And
    Minh Ngoc Dinh\\
    RMIT University, Australia\\
    minh.dinh4@rmit.edu.vn
}

\begin{document}
\maketitle
\begin{abstract}
This paper presents our proposed approach to the Discharge Me! shared task, collocated with the 23th Workshop on Biomedical Natural Language Processing (BioNLP).
In this work, we develop an LLM-based framework for solving the Discharge Summary Documentation (DSD) task, i.e., generating the two critical target sections `Brief Hospital Course' and `Discharge Instructions' in the discharge summary.
By streamlining the recent instruction-finetuning process on LLMs, we explore 
several prompting strategies for optimally adapting LLMs to 
specific generation task of DSD.
Experimental results show that 
providing a clear output structure, complimented by a set of comprehensive Chain-of-Thoughts (CoT) questions,
effectively improves the model's reasoning capability, 
and thereby, enhancing the structural correctness and faithfulness of clinical information in the generated text.
Source code is available at: 
\url{https://github.com/antangrocket1312/Discharge_LLM}
\end{abstract}

\section{Introduction}
Discharge summaries encapsulate key details of a patient's hospitalization, from admission to discharge.
These documents, however, 
can contain excessive amount of medical notes,
making it difficult for subsequent caregivers or patients to quickly understand essential past medical information.
\emph{Brief Hospital Course} and \emph{Discharge Instructions} then become two critical sections in discharge summaries to address this issue.
The former outlines critical hospital events for healthcare providers, while the latter offers post-discharge care instructions to patients and their caregivers. 
The Discharge Me! shared task~\footnote{\url{https://www.codabench.org/competitions/2008/}}~\citep{xu-etal-2024-overview} at the BioNLP Workshop, 
known as 
Discharge Summary Documentation (DSD), focuses on efficiently generating these two critical sections, a task that is both challenging and time-consuming for clinicians.

\begin{figure*}[tbh]
  \centering
  \includegraphics[width=0.8\textwidth]{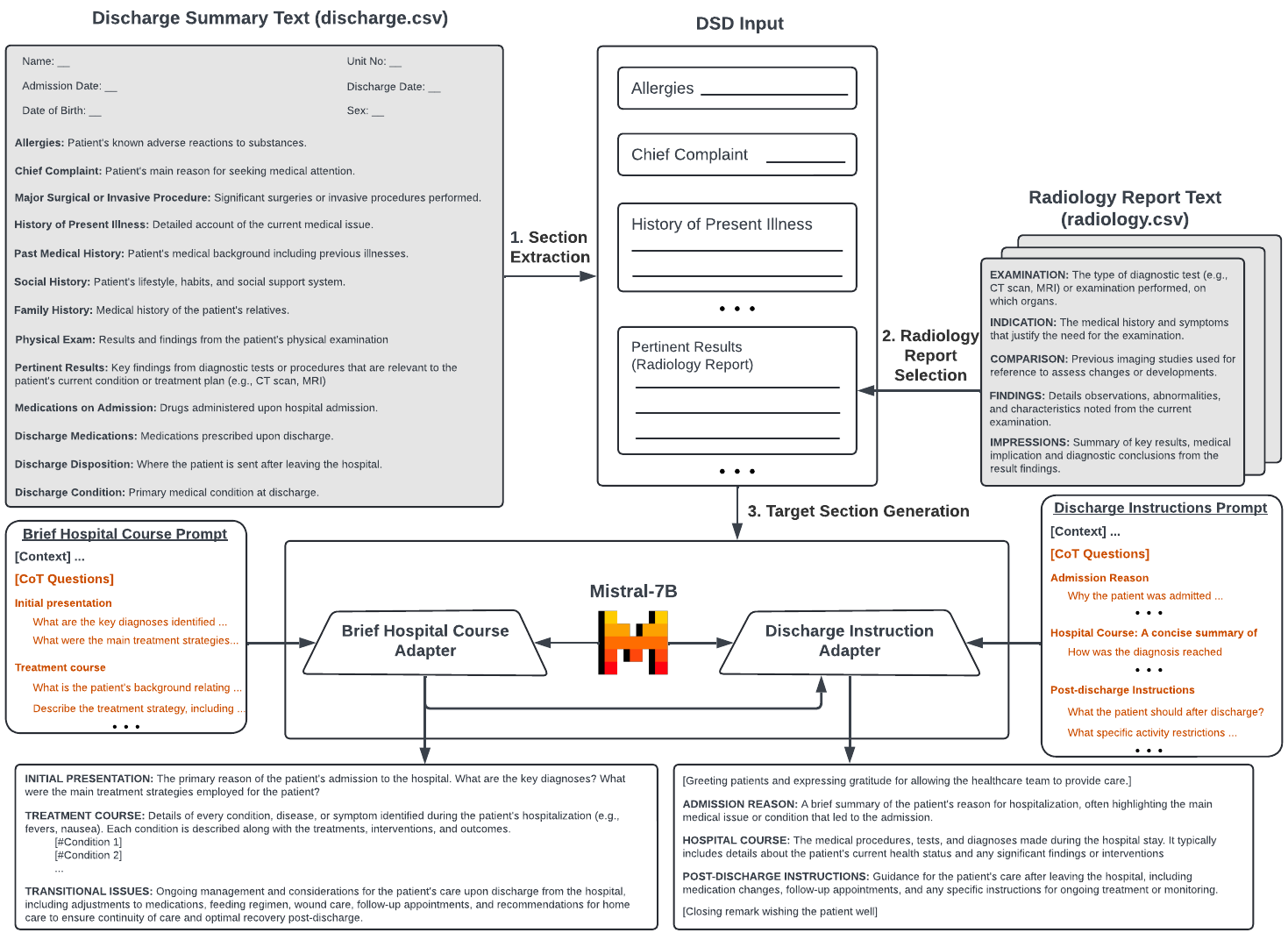}
  \caption{The Discharge-LLM framework}
  \label{fig:Discharge_LLM_Framework}
  \vspace{-1em}
\end{figure*}

In this paper, we introduce a novel LLM-based framework, namely Discharge-LLM, for the DSD task~\citep{xu-etal-2024-overview}.
Discharge-LLM employs modern prompting strategies (e.g., Chain-of-Thought (CoT)) into instruction-finetuning a \texttt{Mistral} Large Language Model (LLM),
which enhances structural correctness and 
faithfulness of clinical information to generate 
the Brief Hospital Course and Discharge Instructions sections of discharge summaries. 

\section{Related Work}
In recent years, Large Language Models (LLMs) like GPT-2 or GPT-3 have excelled in NLP tasks such as language generation, question answering, 
due to their vast number of paramters and extensive training on diverse datasets.
These models can be adapted to new domains and tasks through methods like prompting, which uses natural language instructions~\citep{liu2023prompt} or few-shot examples~\citep{lampinen-etal-2022-language}. 
However, considering DSD problem, the length and excessive information in discharge summaries hinders their use as examples for few-shot prompting.
Alternatively, parameter-efficient fine-tuning, 
which freezes an LLM weights and inserts a small number of tunable parameters~\citep{lin-etal-2020-exploring},
has proven effective in 
specialized clinical tasks like radiology report generation~\citep{van-veen-etal-2023-radadapt}. 

From the clinical summarization perspective, 
research towards th DSD task was very limited.
But there has been a growing focus on many clinical text generation tasks, 
encompassing 
radiology reports~\citep{ben-abacha-etal-2021-overview}, clinical notes~\citep{grambow-etal-2022-domain}, and summary of diagnoses and patient problems~\citep{gao2022summarizing}.

\section{Problem description}
A discharge summary can contain several free-text sections and medical notes compiled from EHR.
The task is to generate the \emph{Brief Hospital Course} (BHC) and \emph{Discharge Instructions} (DI) sections, leveraging readily available data in other sections of discharge summaries and additional information about a patient's admission (e.g.,radiology, stays) from the dataset.
Brief Hospital Course outlines critical hospital events for healthcare providers, while Discharge Instructions offers post-discharge care instructions to patients and their caregivers.

\section{Methodology}
We propose the Discharge-LLM framework, which adapt LLM to the each generation task of DSD, illustrated in Figure~\ref{fig:Discharge_LLM_Framework}.
Discharge-LLM applies three steps, namely \emph{Section Extraction}, \emph{Radiology Report Selection} and \emph{Target Section Generation} to generate the two critical target sections given 
discharge summary and radiology report information of a patient's hospital visit.
Note that we utilize the generated BHC for the subsequent generation of DI.
Table~\ref{table:brief_hospital_course_output} and~\ref{table:discharge_instructions_output} (Appendix~\ref{sec:generation_output}) 
show the output of the two target sections generated by our framework.

\subsection{Section Extraction}
\label{sec:section_parsing}
To generate the two target sections 
BHC and DI, the most straightforward approach is to leverage the other readily available free-text sections in the discharge summary as the input for the generation stage.
But using them all for one-stage generation is overwhelming and prone to hallucination because 
some sections are irrelevant or 
contain thousand words of nonessential information, making key aspects of the patient's record often be omitted.
We thereby design sets of heuristics (e.g., regular expressions) to selectively extract clinical notes information from 13 relevant sections of the discharge summaries (e.g, \emph{History of Present Illness}, \emph{Pertinent Results}, \dots),
with definition of each section described in Figure~\ref{fig:Discharge_LLM_Framework}.
We report 
data distribution
of these sections in Table~\ref{table:discharge_sections_distribution} (Appendix~\ref{sec:discharge_sections_distribution})

\subsection{Radiology Report Selection}
Through exploring the format of different sections of the discharge summary, we notice a great complications in the structure and content of the \emph{Pertinent Results} section, 
likely due to note bloat and information overload.
This section, intended to highlight key findings of radiologist to the patient's treatment, is often cluttered with excessive laboratory and imaging data (e.g., blood tests, CT scans).
These extraneous details can lead to challenges such as hallucination and high resource demands in generative tasks.
Consequently, we explored using radiology reports as a viable alternative. These reports, often duplicated partially or entirely in the Pertinent Results section, succinctly convey diagnoses corresponding to specific lab results. We selected radiology reports with similar Impressions to those in the Pertinent Results and used these as a substitute, streamlining the content effectively.

\subsection{Target Section Generation}
\label{sec:target_section_generation}
In this framework, we performed instruction-finetuning on LLM to adapt the model to DSD.
For computational feasiblity, we employed Low-Rank Adaptation (LoRA)~\citep{hu2021lora}, a parameter-efficient fine-tuning method that adds a small number of trainable parameters to the model while freezing the model's original weight, resulting in standalone adapters.
The adapters, specifically fine-tuned for each generation task in DSD, adjust important weight of LLMs to capture 
 and generate clinical information in the corresponding form.

\paragraph{Prompting Strategies}
Following OpenAI's prompt engineering guidelines~\footnote{\url{https://platform.openai.com/docs/guides/prompt-engineering}}, we structured our prompts into five parts, detailed in Table~\ref{table:prompt_table} (Appendix~\ref{sec:prompt_details}):
1) Context of the discharge summary input to be summarized 2) Definition of the generation task and the specific section for documenting the discharge summary 3) Structure of the expected output of the generating section, infused with 4) Set of Chain-of-Thought (CoT) questions expected to be answered by the LLMs to capture and generate the information in each subsection of the output.
Of those, our primary strategy is Part 5, 
which involved curating effective and generalizable CoT questions based on analysis of numerous samples.
This manual effort helped in designing templates and questions that effectively guide the LLMs to focus on critical information amidst the extensive data and noise in the discharge summaries.
We 
analyzed
the medical questionnaire essential for each section, based on hundreds of samples, in Appendix \ref{sec:target_section_information}
, which underpins our CoT questions and prompt design.

\section{Experiments}
\begin{table*} []
  \centering
  \small
  \begin{subtable}[c]{1\textwidth}
      \centering
      \begin{tabularx}{1\textwidth}{|l|X|X|X|X|l|X|l|l|}
        \hline
            Framework & R-1 & R-2 & R-L & BLEU & BERTScore & Meteor & AlignScore & MEDCON \\ \hline
            \emph{${\rm Discharge\_LLM_{CoT}}$} & \textbf{0.283} & 0.087 & 0.170 & \textbf{0.062} & \textbf{0.368} & \textbf{0.206} & 0.230 & \textbf{0.408} \\
            \emph{${\rm Discharge\_LLM_{Context}}$} & 0.263 & \textbf{0.091} & \textbf{0.178} & 0.058 & 0.365 & 0.191 & \textbf{0.234} & 0.397 \\
            \emph{${\rm Discharge\_LLM_{Base}}$} & 0.240 & 0.074 & 0.159 & 0.043 & 0.347 & 0.170 & 0.221 & 0.376 \\ \hline
        \end{tabularx}
      \caption{Brief Hospital Course Generation \label{table:brief_hospital_course_performance}}
      \vspace{1mm}
  \end{subtable}
  \\
  \begin{subtable}[c]{1\textwidth}
    \centering
    \begin{tabularx}{1\textwidth}{|l|X|X|X|X|l|X|l|l|}
      \hline
          Framework & R-1 & R-2 & R-L & BLEU & BERTScore & Meteor & AlignScore & MEDCON \\ \hline
          \emph{${\rm Discharge\_LLM_{CoT}}$} & \textbf{0.392} & \textbf{0.151} & \textbf{0.246} & \textbf{0.077} & \textbf{0.373} & \textbf{0.272} & \textbf{0.288} & \textbf{0.452} \\
          \emph{${\rm Discharge\_LLM_{Context}}$} & 0.356 & 0.103 & 0.205 & 0.075 & 0.360 & 0.272 & 0.286 & 0.429 \\
          \emph{${\rm Discharge\_LLM_{Base}}$} & 0.335 & 0.102 & 0.215 & 0.041 & 0.324 & 0.181 & 0.251 & 0.318 \\ \hline
      \end{tabularx}
    \caption{Discharge Instructions Generation \label{table:discharge_instructions_performance}}
  \end{subtable}
  \vspace{-2mm}
  \caption{\label{table:prompt_evaluation}Evaluation of prompt variants for finetuning Discharge-LLM}
\end{table*}

\begin{table*} []
  \centering
  \small
  \begin{tabularx}{1\textwidth}{|l|X|X|X|X|l|X|l|l|l|}
    \hline
        Framework & R-1 & R-2 & R-L & BLEU & BERTScore & Meteor & AlignScore & MEDCON & \textbf{Overall} \\ \hline
        \emph{${\rm Discharge\_LLM_{CoT}}$} & 0.370 & 0.131 & 0.245 & 0.068 & 0.360 & 0.314 & 0.215 & 0.324 & \textbf{0.253} \\
        Best ranked system & 0.453 & 0.201 & 0.308 & 0.124 & 0.438 & 0.403 & 0.315 & 0.411 & \textbf{0.332} \\\hline
    \end{tabularx}
  \caption{\label{table:leaderboard} Overall performance on two target section from the shared task's phase 2 leaderboard}
  \vspace{-1em}
\end{table*}
\begin{table}[!ht]
  \centering
  \small
  \begin{tabularx}{0.5\textwidth}{|l|X|X|X|X|}
  \hline
  Target Section & $min$ & $median$ & $mean$ & $max$ \\ \hline
      BHC & 22 & 367 & 425 & 2439 \\ \hline
      DI & 10 & 153 & 201 & 2900 
  \\ \hline
  \end{tabularx}
  \caption{\label{table:word_dist_stat}Statistics of reference text's word count on phase 2's test set}
  \vspace{-1.5em}
\end{table}

\subsection{Baseline and Implementation Details}

To showcase the utility of prompt designing 
for adaptation to the DSD task,
we 
developed three baselines, corresponding to three prompt variants for instruction-finetuning LLMs.
${\rm Discharge\_LLM_{Base}}$ was fine-tuned with no instruction, but only the discharge summaries as input and the respective target section as output.
${\rm Discharge\_LLM_{Context}}$ was fine-tuned with additional natural langauge instructions as prefix to the  discharge summary to provide the context and definition of the task's input/output.
Finally, ${\rm Discharge\_LLM_{CoT}}$ was fine-tuned using prompts 
outlining the structure of the respective generating target section.
Along the structure, we embed some CoT questions to elicit LLMs to generate output aligned with the questions.

We choose \texttt{Mistral}~\footnote{https://mistral.ai/}~\citep{jiang2023mistral} as our LM.
The LLM was fine-tuned on a NVIDIA RTX 4090 GPU, and took 10 hours for fine-tuning each generation task.
The following hyperparameters were used: 1 sample per device, 
a LoRA rank and alpha of 128 and 64 for parameter-efficient fine-tuning, a learning rate of $2e10-4$.
We keep other hyperparameters to their default values.
\paragraph{Metrics}
We followed the organizers to measure textual similarity and factual correctness of the generated text based on several metrics,
including BLEU-4~\citep{papineni-etal-2002-bleu}, ROUGE~\citep{lin-2004-rouge}, BERTScore~\citep{zhang2019bertscore}, Meteor~\citep{banerjee-lavie-2005-meteor}, AlignScore~\citep{zha-etal-2023-alignscore}, and MEDCON~\citep{yim2023aci}.

\paragraph{Dataset}
The dataset for this task was sourced from the MIMIC-IV~\citep{johnson2023mimic} dataset, including 109,168 emergency department (ED) admissions and were split into a training (68,785), a validation (14,719), a phase I testing (14,702), and a phase II testing (10,962) subsets.

\paragraph{Data Preprocessing}
To address the variation in discharge summary length, we select data within the interquartile range (Q1-Q3) for training and validation. 
We further ensure consistency by selecting only samples with discharge summaries containing all 13 common sections and their target sections follow the most common format, as outlined in Figure~\ref{fig:Discharge_LLM_Framework}.
Overall, 11.1k and 8.7k samples were selected for training of 
BHC
and 
DI generation, respectively.
For experiment, due to runtime and computational limitations, we sample 250 hidden entries from each phase's testing data, totaling 500 samples 
for evaluation of each generation task.

\subsection{Results}
Table~\ref{table:prompt_evaluation} presents the performance of models fine-tuned by different prompt variants.
Overall, in both generation tasks, natural language instructions plays a critical role in 
guiding the LLM with comprehensive knowledge to understand the task.
Providing well-described context of the generation task already helps the model 
achieves up to 14\% of performance gain across the metrics and tasks.
Further, infusing CoT questions into the instructions effectively elicit LLM to think better, 
providing an additional 9\% performance increase.
Notably, a reasonable improvement on MEDCON score also indicates better accuracy and consistency of clinical concepts in the generated text.

\subsection{Shared Task's Evaluation Results}
\begin{figure}[t]
  \centering
  \subfloat[Brief Hospital Course]{%
    \includegraphics[width=0.5\textwidth]{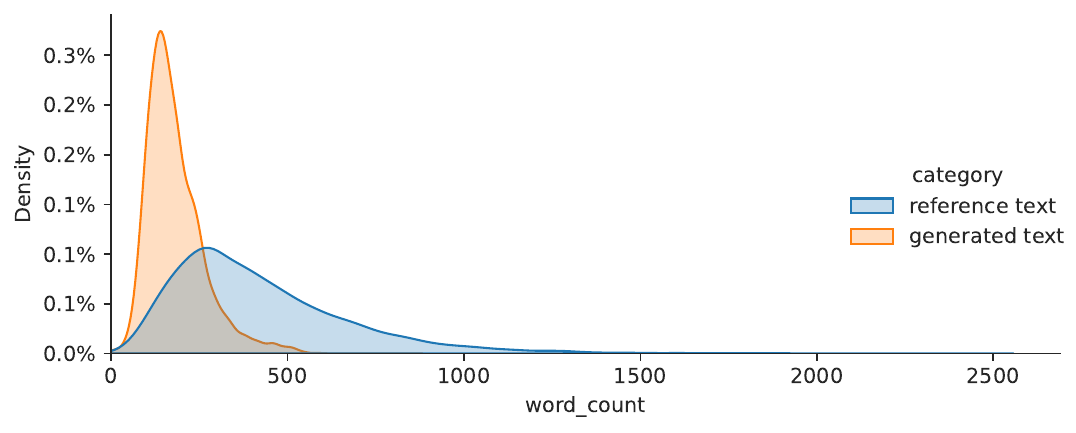}
    \label{fig:evaluation:revenue}%
  }
  \\
  \subfloat[Discharge Instructions]{%
    \includegraphics[width=0.5\textwidth]{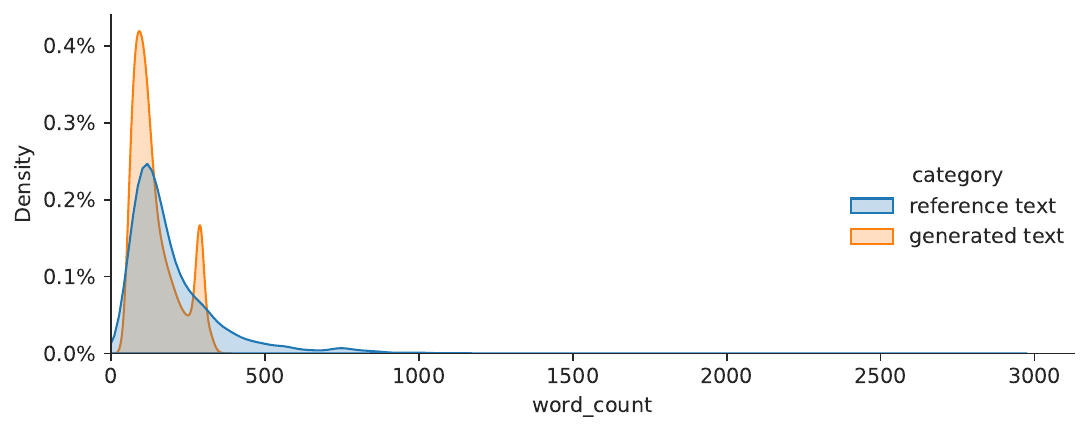}
    \label{fig:evaluation:avgPrice}%
  }
  \caption{\label{fig:word_dist} Distribution of samples per number of words on phase 2's test set}
  \vspace{-1.6em}
\end{figure}

Table~\ref{table:leaderboard} summarizes our framework's overall performance on the phase 2 test set of the shared task, alongside the best-ranked system~\footnote{\url{https://www.codabench.org/competitions/2008/}}.
We notice there is still a gap between our ${\rm Discharge\_LLM_{CoT}}$ framework and the best ranked system, of which the Overall score 
is 0.332.
This performance dip is common across submissions, likely due to prevalent data quality issues in the Discharge Summary Documentation (DSD) task.
DSD, a real-world summarization challenge, involves processing actual 
discharge summary information
with significant variability in formatting and length.
Figure ~\ref{fig:word_dist} shows word distribution variances in the target sections.
It is noticeable our models are trying to set a common length for the target sections, and are struggling to converge to the wide range of lengths of the reference text,
highlighted by Table~\ref{table:word_dist_stat}.
We note slight performance variation of ${\rm Discharge\_LLM_{CoT}}$ in terms of Meteor, AlignScore and MEDCON, in between Table~\ref{table:prompt_evaluation} and~\ref{table:leaderboard}.
A possible reason for such variation may be that hidden test data in the shared task has distributions significantly deviate from the publicly released data for model development.
Furthermore, due to computational constraints and a focus on high-quality data, we utilized only a subset of the available training data. With more comprehensive training, we anticipate improved model convergence.

\section{Conclusion}
In this paper, we present a LLM-based framework for Discharge Summary Documentation that adopts several prompting strategies into instruction-finetuning an LLM,
which enhances structural correctness and 
faithfulness of clinical information in generated target sections.
Using small and open-source LLMs, 
our work also shows the feasiblity of developing and deploying future lightweight NLP systems locally for confidential clinical tasks.
\bibliography{anthology,custom}

\appendix

\section{Generated Output of Brief Hospital Course and Discharge Instructions}
\label{sec:generation_output}
This section presents details of Table~\ref{table:brief_hospital_course_output} and~\ref{table:discharge_instructions_output}, which show the output of the two ``Brief Hospital Course'' and ``Discharge Instructions'' target sections generated by our framework, taken from medical information of patient with $hadm\_id = 21720538$ from the phase 2's test set.

\begin{table*}[!ht]
  \centering
  \begin{tabularx}{0.7\textwidth}{|X|}
      \toprule
      \makecell{
\begin{lstlisting}[frame=none]
___ with PMH of NASH vs cryptogenic cirrhosis, HTN, 
hyperlipidemia, hypothyroidism who presents from ___ 
___ with DVT and renal failure.

# DVT: Patient with extensive RLE DVT extending from the right 
calf veins into the popliteal vein and femoral vein. Patient 
was started on heparin gtt. He was evaluated by ___ who 
recommended discharge home with home ___.

# Renal Failure: Patient presented with Cr 8.9, likely 
pre-renal in etiology. Patient was given albumin and 
diuresis. Cr improved to 3.5 on discharge.

# Cirrhosis: Patient with cirrhosis secondary to NASH vs 
cryptogenic cirrhosis. Patient was continued on lactulose and 
rifaximin. He was also continued on nadolol.

# Hepatic Encephalopathy: Patient with history of hepatic 
encephalopathy. Patient was continued on lactulose and 
rifaximin.

# Ascites: Patient with large volume ascites. Patient was 
continued on diuresis.

# HTN: Patient was continued on home nadolol.

# Hypothyroidism: Patient was continued on home levothyroxine.

# GERD: Patient was continued on home omeprazole.

# Osteopenia: Patient was continued on home calcium and vitamin 
D.

# Allergic Rhinitis: Patient was continued on home cetirizine.

# Chronic Kidney Disease: Patient with chronic kidney disease 
secondary to cirrhosis. Patient was continued on home 
albumin/dietary management.

# Code: DNR/DNI
# Emergency Contact: ___ (daughter) ___

TRANSITIONAL ISSUES:
- Patient will need to follow up with hepatology and ___ 
clinic.
- Patient will need to follow up with ___ clinic for 
continued management of his DVT.
- Patient will need to follow up with ___ clinic for 
continued management of his ascites.
- Patient will need to follow up with ___ clinic for 
continued management of his hepatic encephalopathy.
- Patient will need to follow up with ___ clinic for 
continued management of his cirrhosis.
- Patient will need to follow up with ___ clinic for 
continued management of his hypertension.
- Patient will need to follow up with ___ clinic for 
continued management of his hyperlipidemia.
\end{lstlisting}}\\
      \bottomrule
  \end{tabularx}
  \caption{\label{table:brief_hospital_course_output}
  Generated Output of Brief Hospital Course.}
\end{table*}

\begin{table*}[!ht]
  \centering
  \begin{tabularx}{0.7\textwidth}{|X|}
      \toprule
      \makecell{
\begin{lstlisting}[frame=none]
Dear Mr. ___,

It was a pleasure taking care of you at ___ 
___. You were admitted to the hospital because you 
were having kidney failure. You were treated with medications 
to help your kidneys recover. You also had a blood clot in your 
legs. You were started on blood thinners to help prevent 
further clots. You were also found to have a lot of fluid in your 
belly. You were given medications to help remove the fluid. 

You were also found to have a lot of confusion. You were given 
medications to help with this. You were also found to have a 
blood infection. You were treated with antibiotics. You were 
discharged home with hospice care.

We wish you the best.

Sincerely,
Your ___ Team
\end{lstlisting}}\\
      \bottomrule
  \end{tabularx}
  \caption{\label{table:discharge_instructions_output}
  Generated Output of Discharge Instructions.}
\end{table*}

\section{Data Distribution of Discharge Summary Sections}
\label{sec:discharge_sections_distribution}

\begin{table*}[!ht]
  \centering
  \begin{tabular}{|l|l|l|l|l|}
  \hline
      \emph{Section} & \textbf{train} & \textbf{valid} & \textbf{test (phase 1)} & \textbf{test (phase 2)} \\ \hline
      \emph{Allergies} & 0.999941669 & 0.999795571 & 0.999864232 & 0.999453801 \\ \hline
      \emph{Chief Complaint} & 0.999956252 & 0.999863714 & 1 & 0.9997269 \\ \hline
      \emph{Major Surgical or Invasive Procedure} & 0.518607636 & 0.516456559 & 0.517615912 & 0.517979062 \\ \hline
      \emph{History of Present Illness} & 0.980386152 & 0.982010221 & 0.980788813 & 0.97997269 \\ \hline
      \emph{Past Medical History} & 0.960203576 & 0.96197615 & 0.958387075 & 0.960491579 \\ \hline
      \emph{Social History} & 0.97414472 & 0.976149915 & 0.974204059 & 0.973782431 \\ \hline
      \emph{Family History} & 0.967567883 & 0.968245315 & 0.968094495 & 0.966317706 \\ \hline
      \emph{Physical Exam} & 0.978315397 & 0.980102215 & 0.978141335 & 0.977878926 \\ \hline
      \emph{Pertinent Results} & 0.981231954 & 0.98153322 & 0.981942842 & 0.981429222 \\ \hline
      \emph{Brief Hospital Course} & 1 & 1 & 1 & 1 \\ \hline
      \emph{Medications on Admission} & 0.939787675 & 0.939625213 & 0.937750322 & 0.939007738 \\ \hline
      \emph{Discharge Medications} & 0.980590311 & 0.981192504 & 0.980585161 & 0.981975421 \\ \hline
      \emph{Discharge Disposition} & 0.989121241 & 0.987189097 & 0.987984522 & 0.987528448 \\ \hline
      \emph{Discharge Diagnosis} & 0.991950302 & 0.992231687 & 0.992261218 & 0.993172508 \\ \hline
      \emph{Discharge Condition} & 0.999970834 & 0.999931857 & 1 & 1 \\ \hline
      \emph{Discharge Instructions} & 1 & 1 & 1 & 1 \\ \hline
  \end{tabular}
  \caption{\label{table:discharge_sections_distribution}
  Data Distribution of sections in the discharge summaries in the provided dataset}
\end{table*}

Table~\ref{table:discharge_sections_distribution} presents the percentage distribution of common sections in the discharge summary text of the training, validation and testing subsets.

\section{Questionnaire for Discharge Summary Documentation}
\label{sec:target_section_information}
\subsection{Brief Hospital Course}
\begin{itemize}
  \item \emph{Patient Background and Presenting Complaint}: "What is the patient's background including pre-existing medical conditions, and what symptoms or events led to their current hospital admission?"
  \item Key Diagnoses and Evaluations: "What are the key diagnoses identified during the hospital stay? For each, how was the diagnosis reached, including any significant tests or evaluations conducted?"
  \item Treatment and Management Strategies: "What were the main treatment strategies employed for the patient's conditions during their stay? Include medications adjusted, procedures performed, and any therapeutic interventions."
  \item Complications and Additional Diagnoses: "Were there any complications or additional diagnoses during the hospital stay? How were these addressed and managed?"
  \item Progress and Monitoring: "How did the patient's condition progress throughout the hospital stay, including any monitoring of symptoms, response to treatments, and adjustments made to the treatment plan?"
  \item Support and Consultation Services: "Which specialist services or support consultations were involved in the patientâ€™s care? How did these consultations impact the patientâ€™s treatment plan and recovery?"
  \item Discharge Planning and Instructions: "What were the conditions and considerations for the patientâ€™s discharge? Include the discharge medications, any changes from previous medication regimens, and follow-up care or lifestyle recommendations."
  \item Follow-Up and Post-Discharge Care: "What are the specific follow-up care instructions and any scheduled tests or consultations? Highlight the importance of follow-up for managing ongoing conditions or monitoring recovery."  
\end{itemize}

\subsection{Discharge Instructions}
\begin{itemize}
  \item Initial Assessment and Diagnosis: What led to the patient's admission to the hospital, and what were the initial symptoms? Based on the patient's symptoms, what diagnoses were considered and which was confirmed?
  \item Treatment and Hospital Stay: What treatments were provided to address the patient's symptoms or condition during the hospital stay? Were any surgeries recommended or performed? If a surgery was recommended but not performed, what were the reasons? What were the outcomes of the treatments or interventions provided?
  \item Patient's Decisions and Care Preferences: Did the patient make any specific requests regarding their care, such as refusing a treatment or requesting a transfer? How were these handled? How did the patient's decisions affect their treatment plan and discharge process?
  \item Comprehensive Post-Discharge Instructions: What are the general care instructions for the patient after discharge, including diet, activity level, and medication management? Are there any specific symptoms or signs that the patient should monitor for which would require immediate medical attention? How should the patient manage their regular home medications in addition to any new medications prescribed at discharge?
  \item Activity and Lifestyle Recommendations: What specific activity restrictions or recommendations are given to ensure a smooth recovery? (e.g., weight lifting limits, mobility advice) Are there any restrictions on driving or operating machinery, especially if the patient is taking new or continued pain medication?
  \item Follow-up Care and Monitoring: What follow-up appointments or tests are recommended for the patient? With whom should these appointments be made? How should the patient approach symptom management, especially if they experience pain, dehydration, or other concerning symptoms?
  \item Communication with Healthcare Providers: Under what circumstances should the patient immediately contact their healthcare provider or seek emergency care? What is the recommended way for the patient to communicate with their healthcare team (e.g., phone call, hospital return)?
  \item Encouragement and Support: How can we encourage the patient to adhere to their discharge instructions and reassure them about their recovery process? What resources or support systems can we recommend to the patient for additional help or information post-discharge?
\end{itemize}

\section{Prompts for Discharge Summary Documentation}
\label{sec:prompt_details}
We present the prompts for generation of the two critical ``Brief Hospital Course'' and ``Discharge Instructions'' target sections in Table~\ref{table:prompt_table}
\begin{table*}[!ht]
  \centering
  \small
  \begin{tabularx}{1\textwidth}{|X|X|}
      \toprule
      \textbf{Prompt for Brief Hospital Course Generation} & \textbf{Prompt for Discharge Instructions Generation} \\ \hline
      \makecell{
\begin{lstlisting}[frame=none,  aboveskip=0pt,belowskip=0pt, basicstyle=\tiny]
In this task, you are provided with a Discharge Summary delimited by triple quotes.
Discharge Summaries are documents that outline the care a patient received during their hospital stay, including diagnoses, treatments, and follow-up care instructions, prepared at the time of a patient's discharge. 
Discharge Summaries are split into various sections and written under a variety of headings, relating to admission, diagnosis and relevant discharge information. But the provided Discharge summary will be missing the \"Brief Hospital Course\". \"Brief Hospital Course\" is a section of the discharge summaries that outlines the key events of a patient's hospital stay, including the progression from admission to discharge. It is written for the subsequent care providers about the critical aspects of the patient.
You are tasked to generate the missing \"Brief Hospital Course\" section in the discharge summary, based on the information of other sections in the discharge summary.
Brief Hospital Course outlines the key events of a patient's hospital stay, including the progression from admission to discharge. It is written for the subsequent care providers about the critical aspects of the patient

The summary should be written in the following structure, by answering some important questions:
1. Initial presentation: Describe the patient's initial presentation, including the main complaint and relevant history.
    * What were the main treatment strategies employed for the patient's conditions during their stay? Include medications adjusted, procedures performed, and any therapeutic interventions.
    * What are the key diagnoses identified during the hospital stay?
2. Treatment course:
    - For each section header named by "#Condition Name", provide a detailed description of each condition, disease, or symptom of the patient by answering the following questions:
        * What is the patient's background relating to the condition, disease, or symptom
        * Describe the treatment strategy, including any medications given, procedures performed, and dietary adjustments.
        * How was the diagnosis reached, including any significant tests or evaluations conducted?
        * What were the significant medical or surgical interventions during the hospital stay, including any procedures, diagnostic tests (e.g., CT Scan, Imaging, Blood Test, MRI), and changes in medication?
        * Were there any complications or additional diagnoses during the hospital stay? How were these addressed and managed?
        * How did the patient's condition progress throughout the hospital stay, including any monitoring of symptoms, response to treatments, and adjustments made to the treatment plan?
        * What were the conditions and considerations for the patient discharge? Include the discharge medications, any changes from previous medication regimens, and follow-up care or lifestyle recommendations.
3. Transitional issues: Highlight any transitional care issues addressed during the hospital stay, including changes in medication, dietary adjustments, and specific care instructions.
4. Acute/active issues: Detail the management of acute or active issues encountered during the stay, using the provided structure for each condition.
5. Chronic/stable issues: Summarize how chronic conditions were managed during the stay and any adjustments made to long-term management plans.
\end{lstlisting}} & 
      \makecell{
\begin{lstlisting}[frame=none, aboveskip=0pt,belowskip=0pt, basicstyle=\tiny]
In this task, you are provided with a Discharge Summary delimited by triple quotes.
Discharge Summaries are documents that outline the care a patient received during their hospital stay, including diagnoses, treatments, and follow-up care instructions, prepared at the time of a patient's discharge. 
Discharge Summaries are split into various sections and written under a variety of headings, relating to admission, diagnosis and relevant discharge information. But the provided Discharge summary will be missing the \"Discharge Instructions\". \"Discharge Instructions\" is a section of the discharge summaries that summarizes key events of a patient's hospital stay, including the progression from admission to discharge. and provide detailed guidelines to patients (and often their caregivers) upon discharge from a hospital or healthcare facility, outlining how to care for themselves at home.
You are tasked to generate the missing \"Discharge Instructions\" section in the discharge summary, based on the information of other sections in the discharge summary.
Discharge Instructions summarizes key events of a patient's hospital stay, including the progression from admission to discharge. and provide detailed guidelines to patients (and often their caregivers) upon discharge from a hospital or healthcare facility, outlining how to care for themselves at home.

The summary should be written in the following structure, by answering some important questions:
1. Admission Reason: A concise explanation of:
    * Why the patient was admitted, including any specific conditions or symptoms addressed during the stay.
    * What was the patient's diagnosis upon admission to the hospital?
2. Hospital Course: A concise summary of what happened to the patient's at the hospital
    * How was the diagnosis reached, including any significant tests or evaluations conducted (e.g., CT Scan, Imaging, Blood Test, MRI)?
    * What were the significant medical or surgical interventions during the hospital stay, including any procedures, diagnostic tests (e.g., CT Scan, Imaging, Blood Test, MRI), and changes in medication?
    * Describe the treatment strategy, including any medications given, procedures performed, and dietary adjustments.
    * How did the patient respond to the treatment and procedures? Did the patient make any specific requests regarding their care, such as refusing a treatment or requesting a transfer? How were these handled?
    * Were there any complications or notable improvements in the patient's condition during the stay?
    * What were the outcomes of the treatments or interventions provided?
3. Post-Discharge Instructions:
    + Follow-Up Care:
        * What the patient should do after leaving the hospital?
        * What specific activity restrictions or recommendations are given to ensure a smooth recovery? (e.g., weight lifting limits, mobility advice)
        * Are there any restrictions on driving or operating machinery, especially if the patient is taking new or continued pain medication?
        * Instructions on how to continue treatments started in the hospital, such as new medications or therapy.
    + Medications (Optional): 
        * Comprehensive instructions for all prescribed medications, including dosage, timing, and any specific instructions for use. 
        * How should the patient manage their regular home medications in addition to any new medications prescribed at discharge?
    + Monitoring: 
        * Guidelines on any self-monitoring the patient should perform at home, such as weighing themselves, monitoring blood pressure, or blood sugar levels, with instructions on when to contact their healthcare provider.
        * Are there any specific symptoms or signs that the patient should monitor for which would require immediate medical attention? Under what circumstances should the patient immediately contact their healthcare provider or seek emergency care?
\end{lstlisting}} \\
      \bottomrule
  \end{tabularx}
  \caption{\label{table:prompt_table}
  Prompts for ``Brief Hospital Course'' and ``Discharge Instructions'' generation of the Discharge-LLM framework.}
\end{table*}

\end{document}